\documentclass{article}

\usepackage{microtype}
\usepackage{graphicx}
\usepackage{subfigure}
\usepackage{booktabs} 
\usepackage{xurl} 

\usepackage{hyperref}

\usepackage[accepted]{icml2023}

\usepackage{amsmath}
\usepackage{amssymb}
\usepackage{mathtools}
\usepackage{amsthm}
\usepackage{mathrsfs}

\usepackage[capitalize,noabbrev]{cleveref}

\theoremstyle{plain}

\theoremstyle{definition}

\theoremstyle{remark}

\newcommand{\beq}{\begin{equation}}
\newcommand{\eeq}{\end{equation}}
\newcommand{\ga}{\lower.7ex\hbox{$\;\stackrel{\textstyle>}{\sim}\;$}}
\newcommand{\la}{\lower.7ex\hbox{$\;\stackrel{\textstyle<}{\sim}\;$}}
\newcommand{\f}{\mathbf{f}}

\DeclareMathOperator{\Tr}{Tr}
\DeclareMathOperator*{\argmax}{arg\,max}
\DeclareMathOperator*{\argmin}{arg\,min}

\usepackage[textsize=tiny]{todonotes}

\icmltitlerunning{Oracle-Preserving Latent Flows}

\begin{document}

\twocolumn[
\icmltitle{Oracle-Preserving Latent Flows}



\icmlsetsymbol{equal}{*}

\begin{icmlauthorlist}
\icmlauthor{Alexander Roman}{yyy}
\icmlauthor{Roy T. Forestano}{equal,yyy}
\icmlauthor{Konstantin T. Matchev}{equal,yyy}
\icmlauthor{Katia Matcheva}{equal,yyy}
\icmlauthor{Eyup Unlu}{equal,yyy}
\end{icmlauthorlist}

\icmlaffiliation{yyy}{Institute for Fundamental Theory, Physics Department, University of Florida, Gainesville, FL 32611, USA}

\icmlcorrespondingauthor{Konstantin Matchev}{matchev@ufl.edu}

\icmlkeywords{Machine Learning, ICML}

\vskip 0.3in
]



\printAffiliationsAndNotice{\icmlEqualContribution} 

\begin{abstract}
 
We develop a deep learning methodology for the simultaneous discovery of multiple nontrivial continuous symmetries across an entire labelled dataset. The symmetry transformations and the corresponding generators are modeled with fully connected neural networks trained with a specially constructed loss function ensuring the desired symmetry properties. The two new elements in this work are the use of a reduced-dimensionality latent space and the generalization to transformations invariant with respect to high-dimensional oracles. The method is demonstrated with several examples on the MNIST digit dataset.
\end{abstract}

\section{Introduction}
\label{sec:intro}

Symmetries permeate the world around us and can be found at all scales --- from the microscopic description of subatomic particles in the Standard Model to the large scale structure of the Universe. Symmetries are used as a guiding principle in contemporary science \cite{Gross1996}, as well as in our everyday interactions. Therefore, a fundamental task in data science is the discovery, description, and identification of the symmetries present in a given dataset. By Noether's theorem \cite{Noether1918}, the presence of a continuous symmetry in the data implies that there exists a conservation law which is universally applicable and indispensable in understanding the system's behavior and evolution. At the same time, symmetries can be perceived as aesthetically pleasing in the arts and be used to recognize and evaluate the work of schools and individual artists \cite{Barenboim:2021vzh}. There is a rich mathematical tradition of studying symmetries and their underlying group properties which prove interesting in their own right \cite{wigner1959group}.

Applications of machine learning (ML) to the study of symmetries have been pursued by a number of groups in various  contexts. 
Existing work has focused on
how a given symmetry is reflected in a learned representation of the data \cite{Iten1807.10300,Dillon:2021gag} or in the ML architecture itself, e.g., in the embedding layer of a neural network (NN) \cite{Krippendorf:2020gny}. Other proposals attempt to design special ML architectures, e.g., equivariant NNs, which have a desired symmetry property built in from the outset \cite{Butter:2017cot,Kanwar:2020xzo,Bogatskiy:2020tje,Gong:2022lye,Bogatskiy:2022hub,Li:2022xfc,Hao:2022zns} and test their performance \cite{Gruver2210.02984}. When symmetries are incorporated directly into the ML model, it becomes more economical (in terms of learned representations), interpretable, and trainable. This line of work can be extended to discrete (permutation) symmetries as well \cite{Fenton:2020woz,Shmakov:2021qdz}. 


Lately, machine learning techniques are being applied to more formal mathematical questions that traditionally have fallen within the domain of theorists. For instance, a good understanding of the symmetries present in the problem can reveal conserved quantities \cite{2003.04630,Liu:2020omw} or hint at a more fundamental unified picture \cite{Wu_2019}. ML has been used to discover the symmetry of a potential \cite{Krippendorf:2020gny,Barenboim:2021vzh,Craven:2021ems}, to decide whether a given pair of inputs is related by a symmetry or not \cite{Wetzel:2020jan}, to distinguish between scale-invariant and conformal symmetries \cite{Chen:2020dxg}, and to explore the landscape of string theories \cite{He:2017set,Carifio:2017bov,Ruehle:2020jrk}. Recent work made use of Generative Adversarial Networks (GANs) to learn transformations that preserve probability distributions \cite{Desai:2021wbb}. ML applications have also found their way into group theory, which provides the abstract mathematical language of symmetries. For example, recent work used ML to compute tensor products and branching rules of irreducible representations of Lie groups \cite{Chen:2020jjw} and to obtain Lie group generators of a symmetry present in the data \cite{Liu:2021azq, Moskalev2210.04345, 2023arXiv230105638F}.

The main goal of this paper is to design a deep-learning method that mimics the traditional theorist's thinking and is capable of discovering and categorizing the full set of (continuous) symmetries in a given dataset {\em from first principles}, i.e., without any prior assumptions or prejudice. 
Our study complements and extends previous related work in \cite{Krippendorf:2020gny,Barenboim:2021vzh,Liu:2021azq,Craven:2021ems,Moskalev2210.04345,2023arXiv230105638F}. Our procedure is general and does not need a priori knowledge of what potential symmetries might be present in the dataset --- instead, the symmetries are learned from scratch.

\section{Definition of the Problem}
\label{sec:problem}

Our starting point is a labeled dataset containing $m$ samples of $n$ features and $k$ targets:
\beq
\begin{array}{cccccccc}
    x_1^{(1)}, &  x_1^{(2)}, & \ldots , & x_1^{(n)}; & y_1^{(1)},& y_1^{(2)},& \ldots , & y_1^{(k)}\\
    x_2^{(1)}, &  x_2^{(2)}, & \ldots , & x_2^{(n)}; & y_2^{(1)},& y_2^{(2)},& \ldots , & y_2^{(k)}\\
    \vdots     & \vdots      & \vdots   & \vdots     & \vdots  & \vdots & \vdots & \vdots \\
    x_m^{(1)}, &  x_m^{(2)}, & \ldots , & x_m^{(n)}; & y_m^{(1)},& y_m^{(2)},& \ldots , & y_m^{(k)}\\
\end{array}
\label{eq:dataset}
\eeq
We use boldface vector notation 
\beq
{\mathbf x}\equiv \{x^{(1)}, x^{(2)},\ldots, x^{(n)}  \}\in {\mathbb{R}^n}
\eeq
for the $n$-dimensional input features and arrow vector notation for the $k$-dimensional target vectors 
\beq
\vec y \equiv \{y^{(1)}, y^{(2)}, \ldots, y^{(k)} \}\in {\mathbb{R}^k}.
\eeq
The dataset (\ref{eq:dataset}) can then be written in a compact form as $\left\{\mathbf{x}_i; \vec{y}_i \right\}$, where $i=1,2,\ldots,m$.

In order to define a symmetry of the data (\ref{eq:dataset}), we shall utilize the vector function $\vec{\varphi}(\mathbf{x})$ which plays the role of an oracle producing the corresponding target labels $\vec{y}_1,\vec{y}_2,\ldots,\vec{y}_m$:
\beq
\vec{y}_i \; \equiv \; \vec{\varphi}(\mathbf{x}_i) \, , \qquad i \, = \,1,2,\ldots,m \, .
\label{eq:labels}
\eeq
This is an ideal classifier on the given dataset.
In data science applications (our primary interest in this paper), the vector oracle $\vec{\varphi}: \mathbb{R}^n \rightarrow{\mathbb{R}^k}$ needs to be learned numerically from the dataset (\ref{eq:dataset}) via standard regression methods. On the other hand, in certain more formal theoretical applications, the oracle $\vec{\varphi}$ can be already given to us externally in the form of an analytical function. The methodology developed in this paper applies to both of these situations.

With this setup, the main goal will be to derive a symmetry transformation 
$\f: \mathbb{R}^n \rightarrow \mathbb{R}^n$
\beq
\mathbf{x}' \; = \; \f(\mathbf{x})  \, ,
\label{eq:symmetry}
\eeq
which preserves the $\vec{\varphi}$-induced labels (\ref{eq:labels}) of our dataset (\ref{eq:dataset}). In other words, we want to find the function $\f(\mathbf{x})$ for which 
\beq
\vec{\varphi}(\mathbf{x}'_i) \; \equiv \; \vec{\varphi}(\f(\mathbf{x}_i)) = \vec{\varphi}(\mathbf{x}_i), \quad \forall i \,= \,1,2,\ldots,m \, .
\label{eq:symmetrycondition}
\eeq

In general, a given dataset will exhibit several different symmetries, which can now be examined and categorized from the point of view of group theory. For this purpose, one needs to focus on infinitesimal symmetry transformations and study the corresponding set of distinct generators $\{\mathbf{J}_\alpha\}$, $\alpha=1,2,\ldots,N_g$. A given set of generators $\{\mathbf{J}_\alpha\}$ forms a Lie algebra if the closure condition is satisfied, i.e., if all Lie brackets $\bigl[ \, .\, ,\, .\, \bigr]$ can be represented as linear combinations of the generators already present in the set:
\beq
\bigl[ \mathbf{J}_\alpha, \mathbf{J}_\beta\bigr] 
= \sum_{\gamma=1}^{N_g} a_{[\alpha\beta]\gamma} \mathbf{J}_\gamma .
\label{eq:algebraclosure}
\eeq
The coefficients $a_{[\alpha\beta]\gamma}$ are the structure constants of the symmetry group present in our dataset (the square bracket index notation reminds the reader that they are anti-symmetric in their first two indices: $a_{[\alpha\beta]\gamma}=-a_{[\beta\alpha]\gamma}  $). The simplest algebras (referred to as Abelian) are those for which all generators commute and hence have vanishing structure constants. These are the types of algebras which we will encounter in the two toy examples of this paper.

\section{Method}
\label{sec:solution}

In our approach, we model the function $\f$ with a neural network ${\mathcal F}_{\mathcal W} $ with $n$ neurons in the input and output layers, corresponding to the $n$ transformed features of the data point $\mathbf{x}'$. The trainable network parameters (weights and biases) will be generically denoted with $\mathcal W$. During training, they will evolve and converge to the corresponding {\em trained} values $\widehat{\mathcal W}$ of the parameters of the trained network ${\mathcal F}_{\widehat{\mathcal W}}$, i.e., the hat symbol will denote the result of the training. 
In order to ensure the desired properties of the network, we design a loss function with the following elements.

{\bf Invariance.} In order to enforce the invariance under the transformation (\ref{eq:symmetry}), we include the following mean squared error (MSE) term in the loss function $L$:
\beq
    L_\text{inv}(\mathcal W, \{\mathbf x_i; \vec{y}_i\}) = \frac{1}{m}\sum_{i=1}^m \left[ \vec{\varphi}\left({\mathcal F}_{\mathcal W}(\mathbf{x}_i)\right)-\vec{y}_i \right]^2 \, .
\label{eq:lossInv}
\eeq
A NN trained with this loss function will find an arbitrarily general (finite) symmetry transformation ${\mathcal F}_{\widehat{\mathcal W}}$ parametrized by the values of the trained network parameters $\widehat{\mathcal W}$. The particular instantiation of ${\mathcal F}_{\widehat{\mathcal W}}$ 
depends on the initialization of the network parameters, so by repeating the procedure with different initializations one obtains a family of symmetry transformations.

{\bf Infinitesimality.} 
In order to focus on the symmetry {\em generators}, we restrict ourselves to infinitesimal transformations $\delta{\mathcal F}$ in the vicinity of the identity transformation $\mathbf{I}$:
\beq
\delta{\mathcal F} \; \equiv \; \mathbf{I} + \varepsilon \, {\mathcal G}_{\mathcal W} \, ,
\label{eq:fepsilon}
\eeq
where $\varepsilon$ is an infinitesimal parameter and the parameters ${\mathcal W}$ of the new neural network ${\mathcal G}$ will be forced to be finite.
The loss function (\ref{eq:lossInv}) can then be rewritten as
\beq
    L_\text{inv}(\mathcal W, \{\mathbf x_i; \vec{y}_i\}) = \frac{1}{m\varepsilon^2}\sum_{i=1}^m \left[ \vec\varphi(\mathbf{x}_i + \varepsilon {\mathcal G}_{ \mathcal W}(\mathbf{x}_i))-\vec{y}_i \right]^2 \, ,
\label{eq:LossInvInfinitesimal}    
\eeq
with an extra factor of $\varepsilon^2$ in the denominator to compensate for the fact that generic transformations scale as $\varepsilon$ \cite{Craven:2021ems}.
In addition, we add a normalization loss term
\begin{eqnarray}
    &L_\text{norm}(\mathcal W, \{\mathbf x_i\}) = \frac{1}{m}\sum_{i=1}^m \left[ \|{\mathcal G}_{ \mathcal W}(\mathbf{x}_i)\| - 1\right]^2 \nonumber \\[2mm]
    &+~ \frac{1}{m}\sum_{i=1}^m \left[ \|{\mathcal G}_{ \mathcal W}(\mathbf{x}_i)\| - \overline{\|{\mathcal G}_{ \mathcal W}(\mathbf{x}_i)\|}\right]^2.
\label{eq:LossNormInfinitesimal}    
\end{eqnarray}

After minimization of the loss function, the trained NN ${\mathcal G}_{\widehat {\mathcal W}}$ will represent a corresponding generator
\beq
\mathbf{J} \; = \; {\mathcal G}_{\widehat {\mathcal W}} \, ,
\eeq
where
\beq
\widehat{\mathcal W}\equiv\argmin_{\mathcal W}
\biggl(L_\text{inf} + h_\text{norm} L_\text{norm} \biggr)
\eeq
are the learned values of the NN parameters ($h_\text{norm}$ is a hyperparameter usually chosen to be 1).
By repeating the training $N_g$ times under different initial conditions ${\mathcal W}_0$, one obtains a set of $N_g$ (generally distinct) generators $\{\mathbf{J}_\alpha\}$, $\alpha=1,2,\ldots,N_g$.

{\bf Orthogonality.} 
To ensure that the generators $\{\mathbf{J}_\alpha\}$ are distinct, we introduce an additional orthogonality term to the loss function
\beq
 L_\text{ortho}({\mathcal W, \{\mathbf x_i\}}) = 
 \frac{1}{m} \sum_{i=1}^m 
\sum_{ \alpha < \beta}^{N_g} 
\left[
{\mathcal G}_{{\mathcal W}_\alpha}(\mathbf{x}_i)\cdot {\mathcal G}_{{\mathcal W}_\beta}(\mathbf{x}_i)\right]^2. 
\label{eq:lossOrtho}
\eeq

{\bf Group structure.} 
In order to test whether a certain set of {\em distinct} generators $\{\mathbf{J} _\alpha\}$ found in the previous steps generates a group, we need to check the closure of the algebra (\ref{eq:algebraclosure}), e.g., by minimizing
\beq
L_\text{closure} (a_{[\alpha\beta]\gamma}) = \sum_{\alpha<\beta}\Tr \left(\mathbf{C}_{[\alpha\beta]}  ^T\mathbf{C}_{[\alpha\beta]}\right),
\label{eq:LossClosure}
\eeq
with respect to the candidate structure constant parameters $a_{[\alpha\beta]\gamma}$, where the closure mismatch is defined by
\beq 
\mathbf{C}_{[\alpha\beta]} (a_{[\alpha\beta]\gamma}) \equiv
\bigl[ \mathbf{J}_\alpha, \mathbf{J}_\beta\bigr] 
- \sum_{\gamma=1}^{N_g} a_{[\alpha\beta]\gamma} \mathbf{J}_\gamma .
\label{eq:closuremismatch}
\eeq
Since $L_\text{closure}$ is positive semi-definite, $L_\text{closure}=0$ would indicate that the algebra is closed and we are thus dealing with a genuine (sub)group. 


In principle, the number of generators $N_g$ is a hyperparameter that must be specified ahead of time. Therefore, when a closed algebra for a given $N_g$ value is found, it is only guaranteed to be a subalgebra of the full symmetry group, and one must proceed to try out higher values for $N_g$ as well. The full algebra will then correspond to the maximum value of $N_g$ for which a closed algebra of distinct generators is found to exist.

\section{Simulation Setup}
\label{sec:simulation}


In order to illustrate our method, we use the standard  MNIST dataset \cite{lecun-mnisthandwrittendigit-2010} consisting of 60000 images of 28 by 28 pixels. In other words, the dataset (\ref{eq:dataset}) consists of $m=60000$ samples of $n=28\times28=784$ features. The images are labelled as a digit from 0 to 9. The classic classification task for this dataset is to build an oracle $\vec y = \vec{\varphi}(\mathbf x)$ which calculates a ten-dimensional vector of `logits' in the output layer (this raw output is typically normalized into respective probabilities with a softmax function). An image is then classified as $\argmax(\vec{\varphi}(\mathbf x))$.

When applied to the MNIST dataset, the general problem of Section~\ref{sec:problem} can be formulated as follows: what types of transformations (\ref{eq:symmetry}) can be performed on {\em all of} the original images in the dataset so that the 10-dimensional logits vector $\vec{\varphi}(\mathbf x)$ is conserved, i.e., that the symmetry equation (\ref{eq:symmetrycondition}) is satisfied. Note that conserving all ten components of the oracle function as in eq.~(\ref{eq:symmetrycondition}) is a much stronger requirement than simply demanding that the prediction for each image stays the same --- for the latter, it is sufficient to ensure that the map $\f$ is such that
\beq
\argmax(\vec{\varphi}(\f(\mathbf x))) = \argmax(\vec{\varphi}(\mathbf x)).
\label{eq:weakersymmetry}
\eeq
As discussed in Section~\ref{sec:solution} and illustrated with the examples in the next section, we apply deep learning to derive the desired transformations $\f$. 
The neural network ${\mathcal F}_{\mathcal W}$ is implemented as a sequential feed-forward neural network in {\sc PyTorch} \cite{NEURIPS2019_9015}. 
 Optimizations are performed with the {\sc Adam} optimizer with a learning rate between $3\times 10^{-5}$ and $0.03$. The loss functions were designed to achieve a fast and efficient training process without the need for extensive hyperparameter tuning. The training and testing data (\ref{eq:dataset}) was from the standard  MNIST dataset \cite{lecun-mnisthandwrittendigit-2010}.

\subsection{Trivial Symmetries from Ignorable Features}
\label{sec:cyclic}

\begin{figure}[t]
\vskip 0.2in
\begin{center}
\includegraphics[width=0.49\columnwidth]{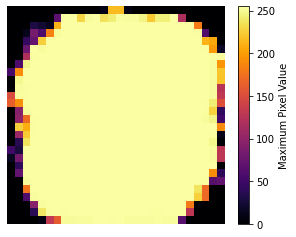}
\includegraphics[width=0.49\columnwidth]{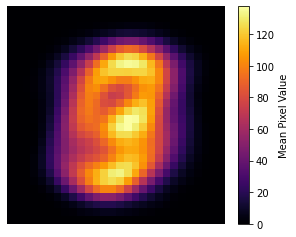}
\caption{A heatmap of the maximum value (left panel) or the mean value (right panel) of each pixel over the MNIST dataset. The individual pixel values in the dataset range from 0 to 255.  }
\label{fig:pixels}
\end{center}
\end{figure}

\begin{figure}[t]
\vskip 0.2in
\begin{center}
\includegraphics[width=0.8\columnwidth]{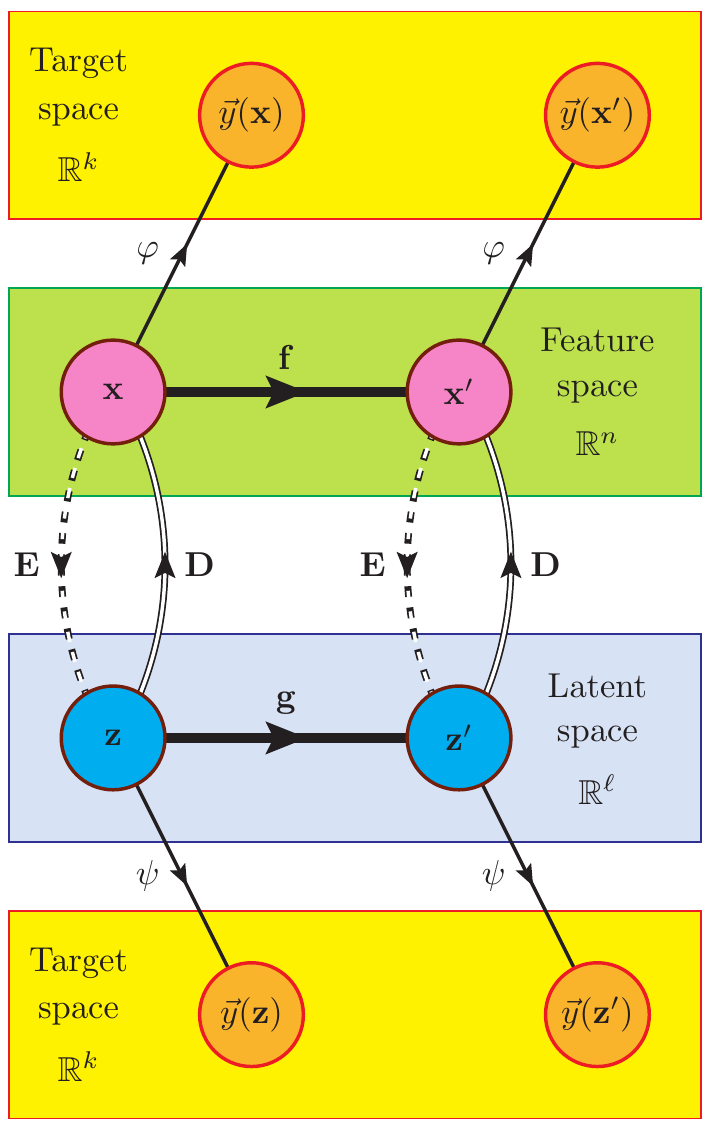}
\caption{Flowchart of the different data processing steps discussed in this paper: an encoder $\mathbf E: \mathbb{R}^n \rightarrow \mathbb{R}^\ell$; a decoder $\mathbf D: \mathbb{R}^\ell \rightarrow \mathbb{R}^n$;
a transformation in the feature space $\f: \mathbb{R}^n \rightarrow \mathbb{R}^n$; 
a transformation in the latent space ${\mathbf g}: \mathbb{R}^\ell \rightarrow \mathbb{R}^\ell$, 
a trained vector oracle $\vec\varphi: \mathbb{R}^n \rightarrow \mathbb{R}^k$, and a trained vector oracle $\vec\psi: \mathbb{R}^\ell \rightarrow \mathbb{R}^k$.
 The transformations $\f$ and ${\mathbf g}$ are symmetries if $\vec{y}(\mathbf x')=\vec{y}(\mathbf x)$ and $\vec{y}(\mathbf z')=\vec{y}(\mathbf z)$, respectively. }
\label{fig:flowchart}
\end{center}
\end{figure}

Before proceeding, let us develop some intuition by discussing the trivial symmetries which are present in the dataset in analogy to the treatment of cyclic (sometimes also called ignorable) coordinates in classical mechanics \cite{goldstein:mechanics}. In classical mechanics, if the Lagrangian of a system does not contain a certain coordinate, then that coordinate is said to be cyclic or ignorable and its corresponding conjugate momentum is conserved. Let us ask whether there are any such ignorable features in our case. The two heatmaps in Figure~\ref{fig:pixels} show the maximum value for each pixel (left panel) as well as the mean value for each pixel (right panel), when averaged over the whole dataset. Figure~\ref{fig:pixels} reveals that there are a number of pixels near the corners and the edges of the image which contain no data at all. 
This suggests that a robust classifier will be insensitive to the values of those pixels, i.e., these pixels behave similarly to the ignorable coordinates seen in classical mechanics. This observation is further evidenced by the fact that the highest, i.e., least informative, principal components across the features (not shown here) are linear combinations of those corner and/or edge pixels. We are not interested in such trivial symmetries. It may be possible to find non-trivial maps directly between images using a deep NN, but the training is predicated on the existence of a metric capable of capturing non-trivial structure in the dataset.\footnote{For example, rotating an image slightly may be a symmetry with respect to a classifier, but it changes the values of many pixels. Hence, the value of a naive metric such as MSE between these two images would be large, on the other hand the distance as measured by a more sophisticated metric such as the Earth Movers Distance would be small.} One candidate for such a metric would be the Earth Movers Distance which solves an optimal transport problem between two distributions, but as such, it is too computationally expensive to be used in the loss function.
In what follows we shall first reduce the dimensionality of the dataset and explore symmetry transformations in the corresponding $\ell$-dimensional latent space ${\mathbb{R}^\ell}$, with $\ell\ll n$. This is because in latent space the Euclidean metric captures rich structure.


\subsection{Dimensionality Reduction}
\label{sec:dimred}

\begin{figure*}[t]
\vskip 0.2in
\begin{center}
\centerline{\includegraphics[width=1.99\columnwidth]{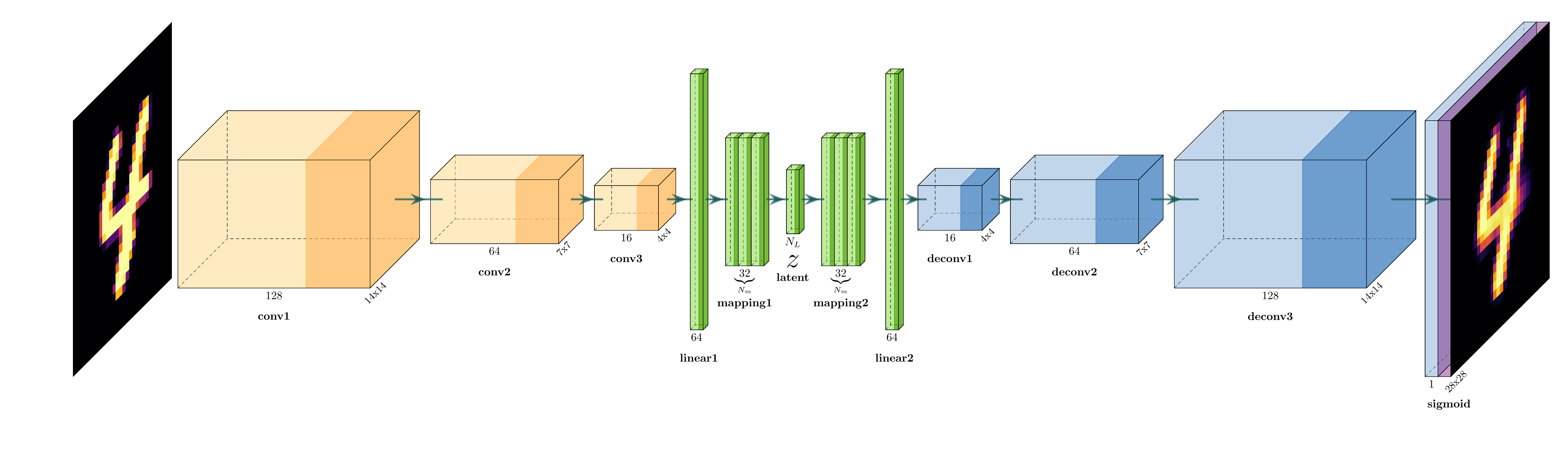}}
\caption{The network architecture of our autoencoder consisting of an encoder $\mathbf E$ and a decoder $\mathbf D$. The yellow modules are convolutional layers, the green modules are fully-connected layers, and the blue modules are convolution-transpose layers. The dark-shaded regions indicate ReLU activation functions.}
\label{fig:autoencoder}
\end{center}
\end{figure*}

As shown in Figure~\ref{fig:flowchart}, instead of looking for transformations $\f$ in the original feature space $\mathbb{R}^n$ that are symmetries that preserve the oracle $\vec\varphi$ from (\ref{eq:labels}), we choose to search for transformations $\mathbf g$ acting in a latent space $\mathbb{R}^\ell$, which are symmetries that preserve the \textit{induced} oracle $\vec\psi: \mathbb{R}^\ell \rightarrow \mathbb{R}^k$. For this purpose, we train an autoencoder whose architecture is shown in Figure~\ref{fig:autoencoder} --- it consists of an encoder $\mathbf E: \mathbb{R}^n \rightarrow \mathbb{R}^\ell$ and a decoder $\mathbf D: \mathbb{R}^\ell \rightarrow \mathbb{R}^n$. The latent vectors $\mathbf{z}_i\in \mathbb{R}^\ell$ are defined as
\begin{eqnarray}
{\mathbf z}_i &\equiv& {\mathbf E}(\mathbf{x}_i) \, , \qquad i \, = \,1,2,\ldots,m \, . \label{eq:encoder}
\end{eqnarray}
The induced classifier $\vec\psi: \mathbb{R}^\ell \rightarrow \mathbb{R}^k$ is trained as
\beq
\vec{y}_i = \vec\psi(\mathbf z_i)
= \vec\psi({\mathbf E}(\mathbf{x}_i)),  \quad i \, = \,1,2,\ldots,m \, , \label{eq:psi}
\eeq
and is built as a fully connected dense NN with three hidden layers of sizes 128, 128 and 32, trained with the categorical cross-entropy loss. 
%

With those tools in hand, we shall look for symmetry transformations 
${\mathbf g}: \mathbb{R}^\ell \rightarrow \mathbb{R}^\ell$ in the latent space in analogy to eq.~(\ref{eq:fepsilon})
\beq
\mathbf{z}' \; = \; \mathbf{z} + \varepsilon\, {\mathbf g}(\mathbf{z})  \, ,
\label{eq:latent_transformation}
\eeq
which preserve the new oracle $\psi$ in analogy to (\ref{eq:symmetrycondition})
\beq
\vec{\psi}(\mathbf{z}'_i) \; \equiv \; \vec{\psi}(\mathbf{z} + \varepsilon\, {\mathbf g}(\mathbf{z})) = \vec{\psi}(\mathbf{z}_i), \quad \forall i \,= \,1,2,\ldots,m \, .
\label{eq:latentsymmetrycondition}
\eeq
A single transformation ${\mathbf g}$ will be represented with a fully connected dense NN with various architectures, as necessitated by the complexity of the exercise, and trained with the loss functions described in Section~\ref{sec:solution}. Once such a symmetry transformation ${\mathbf g}$ in the latent space is found, its effect on the actual images can be illustrated and analyzed with the help of the decoder ${\mathbf D}$.

\section{Examples}
\label{sec:examples}

For the main exercise presented in Section~\ref{sec:symmetries} below, we shall use the complete set of digits and a 16-dimensional latent space, in which, however, the results will be difficult to plot and visualize. This is why we begin with a couple of toy examples, where we consider only two classes, the zeros and the ones, and either a two-dimensional latent space (Section~\ref{sec:2digits2var}) or a three-dimensional latent space (Section~\ref{sec:2digits3var}).

\subsection{Two Categories and $\ell=2$ Latent Variables}
\label{sec:2digits2var}

\begin{figure}[t]
\vskip 0.2in
\begin{center}
\centerline{\includegraphics[width=0.95\columnwidth]{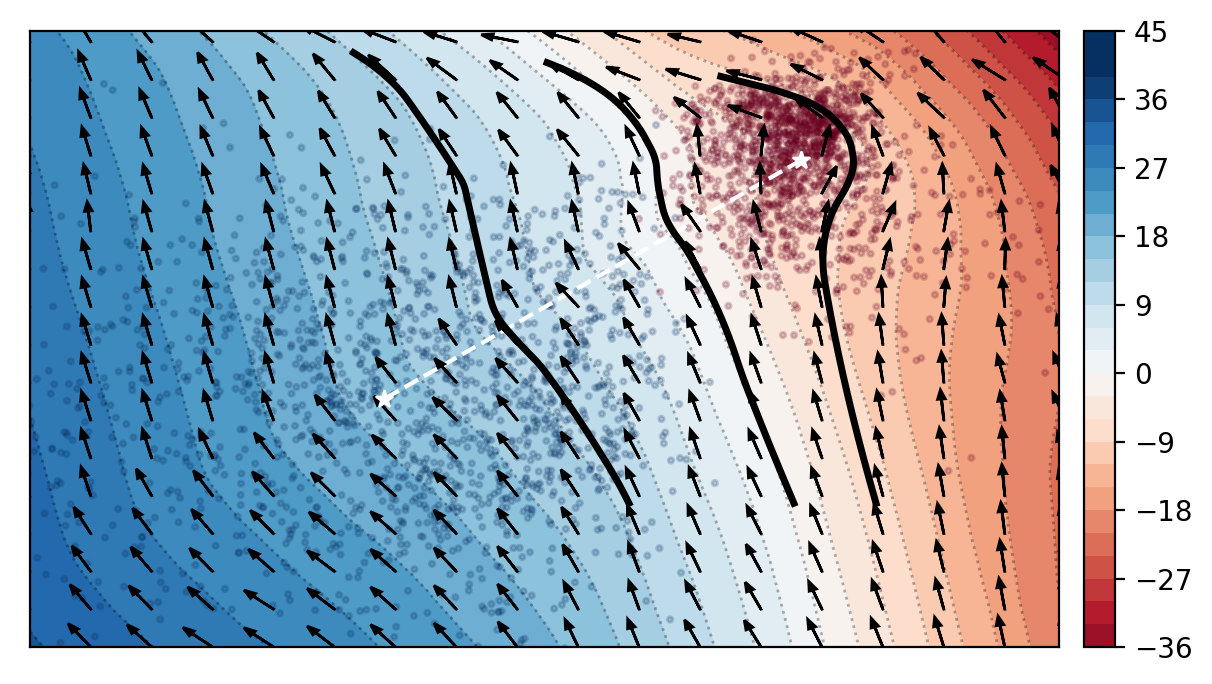}}
\caption{The results from the exercise in Section~\ref{sec:2digits2var} are illustrated in the two-dimensional latent space. The red and blue points represent validation images with true labels 0 and 1, respectively, and the white stars (connected by a straight dashed line) denote the centers of these two clusters. The heatmap shows the unscaled output from the single neuron in the output layer of the $\psi$ classifier. The superimposed vector field visualizes the symmetry transformation $\mathbf g$ found by our method. The black solid lines are three representative symmetry streamlines used for the illustrations in Figure~\ref{fig:2v2d_platonic}.}
\label{fig:2v2d_latent}
\end{center}
\vskip -0.2in
\end{figure}

\begin{figure}[t]
\vskip 0.2in
\begin{center}
\centerline{\includegraphics[width=0.95\columnwidth]{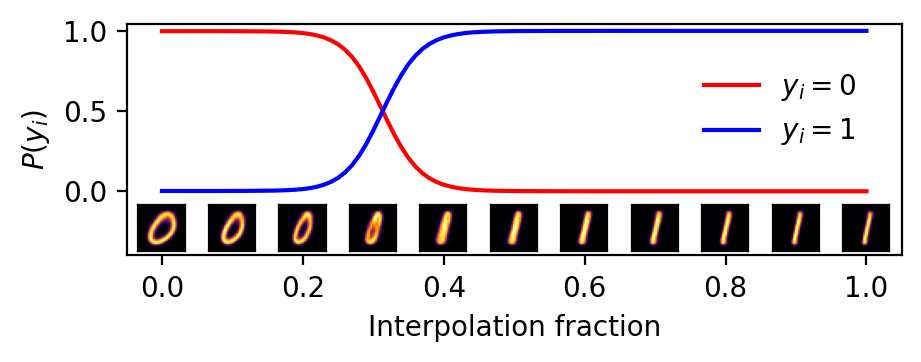}}
\centerline{\includegraphics[width=0.95\columnwidth]{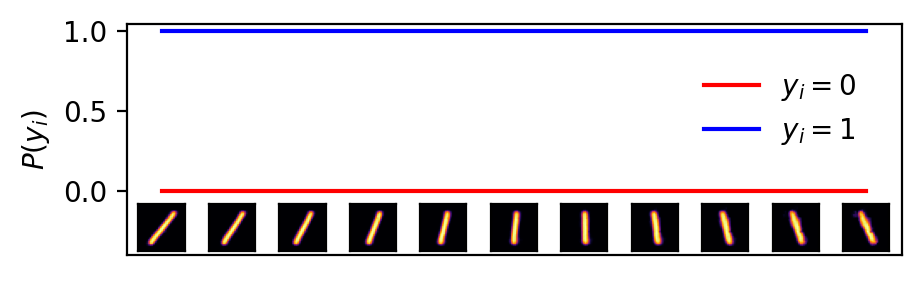}}
\centerline{\includegraphics[width=0.95\columnwidth]{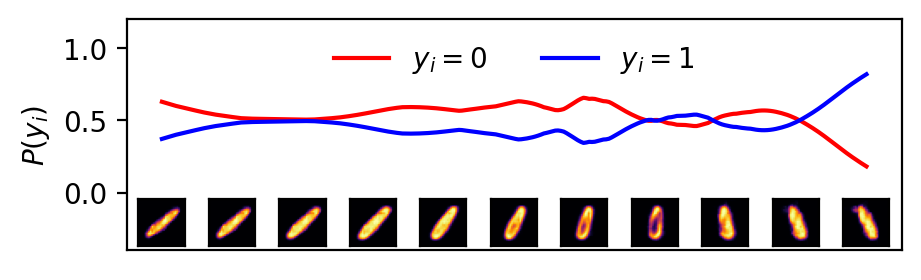}}
\centerline{\includegraphics[width=0.95\columnwidth]{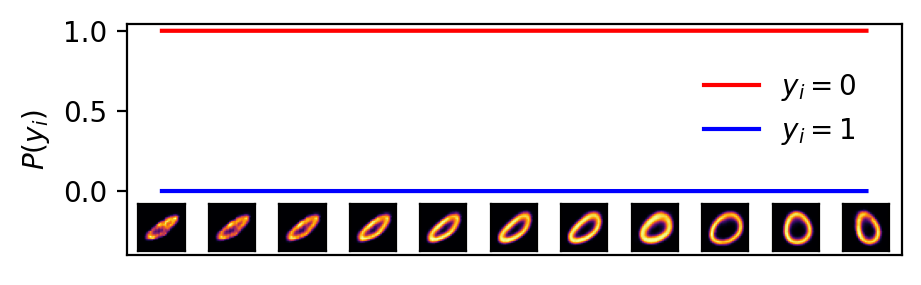}}
\caption{Explorations of the latent space from Figure~\ref{fig:2v2d_latent} by following the white dashed line (top panel), the left solid black line (second panel), the middle solid black line (third panel) or the right solid black line (fourth panel). Each panel shows the likelihood for an image to be zero (red line) or one (blue line) along the respective latent space trajectory. At the base of each panel, we show a row of representative images after applying the decoder ${\mathbf D}(\mathbf z)$.}
\label{fig:2v2d_platonic}
\end{center}
\end{figure}

In this subsection we perform a toy binary classification exercise in $\ell=2$ latent space dimensions. We keep only the images of class ``0" and ``1", which are then randomly train-test split in a ratio of 3:1. The training set is used to train the autoencoder of Figure~\ref{fig:autoencoder} and the classifier $\psi$. Since this example is a simple binary classification, $\psi$ is chosen to have a single output layer neuron, whose logit (raw output) is fed into a sigmoid function. The results are shown in Figure~\ref{fig:2v2d_latent} in the plane of the two latent space variables $(z^{(1)},z^{(2)})$. The red and blue points denote the set of validation images with true labels 0 and 1, respectively. The white star symbols mark the centers of these two clusters, which we refer to as the ``platonic"\footnote{In the sense that they are the ideal representatives of their respective classes.} images of the digits zero and one. By moving along the straight white dashed line between them, we pass through points in the latent space which (after decoding) produce images that smoothly interpolate between the platonic zero and the platonic one. This variation is illustrated in the top panel of Figure~\ref{fig:2v2d_platonic}. Motion along the white dashed line is therefore not a symmetry transformation, since it changes the meaning of the image. It is the motion in the orthogonal direction that we are interested in, since that is the direction in which, while the actual image is changing, its interpretation by the classifier is not.

Once we train the neural network for $\mathbf g$, we obtain a vector field in the latent space. The \textit{flow} of this vector field is illustrated with the black arrows in Figure~\ref{fig:2v2d_latent}. The three black solid lines are three representative symmetry streamlines used for the illustrations in the bottom three panels in Figure~\ref{fig:2v2d_platonic}. The leftmost streamline passes near the platonic one and therefore represents a series of images which are interpreted by the classifier $\psi$ as ``ones" with very high probability (see the blue line in the second panel of Figure~\ref{fig:2v2d_platonic}). The corresponding row of decoded images in the second panel reveals that the symmetry transformation has the effect of rotating the digit ``one" counterclockwise.

\begin{figure}[t]
\vskip 0.2in
\begin{center}
\centerline{\includegraphics[width=0.95\columnwidth]{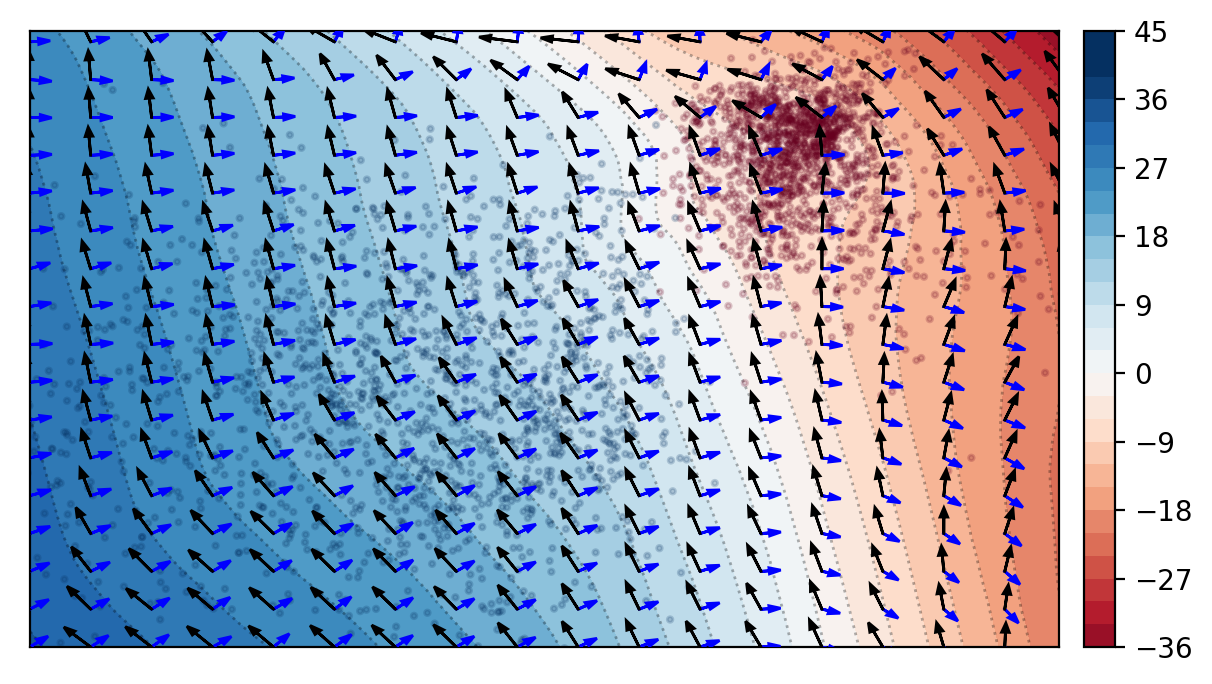}}
\caption{The same as Figure~\ref{fig:2v2d_latent}, but requiring two separate orthogonal symmetry transformations ${\mathbf g}_1$ (black arrows) and ${\mathbf g}_2$ (blue arrows). }
\label{fig:2v2d_latent_2gens}
\end{center}
\end{figure}

The rightmost streamline in Figure~\ref{fig:2v2d_latent}, on the other hand, passes through the region near the platonic ``zero" and therefore contains images which are almost surely interpreted as zeros by the classifier, as evidenced by the red line in the last panel of Figure~\ref{fig:2v2d_platonic}. The corresponding row of decoded images in the last panel reveals that the symmetry transformation has the effect of not only rotating the digit ``zero" counterclockwise, but also simultaneously stretching and enlarging the image.

The middle streamline in Figure~\ref{fig:2v2d_latent} passes through the boundary region between the two category clusters and will therefore generate images which are rather inconclusive according to the classifier. This is confirmed in the third panel of Figure~\ref{fig:2v2d_platonic}, which shows that the likelihoods of a ``zero" and ``one" are comparable\footnote{It is noteworthy that the line is somewhat unstable --- this is because there are very few samples in this region and hence the generator model is less reliable here.} along that streamline. The decoded images are indeed confusing to interpret, even for a human, and are rotated counterclockwise in a similar fashion to the images in the other panels. 

An important feature of our scheme is the ability to find multiple non-trivial symmetries, and vice versa, to determine when no additional symmetries are possible. Since we have chosen a compressed representation with only two dimensions, one of which is constrained by the oracle, this leaves us with a single symmetry degree of freedom. Therefore, if we try to look for a second symmetry (by simultaneously training two NNs for $\mathbf g$, as explained in Section~\ref{sec:solution}) the training will converge to a large loss value, implying that all desired conditions cannot be satisfied simultaneously. Specifically, invariance requires that the flows follow the contours of equal likelihood, thus forcing them to align, which contradicts the orthogonality condition. This dilemma is illustrated in Figure~\ref{fig:2v2d_latent_2gens}, where we repeat the previous exercise for the case of $N_g=2$. The two latent flows, represented with the black and blue arrows respectively, try to strike an optimal balance between these competing and irreconcilable requirements.

\subsection{Two Categories and $\ell=3$ Latent Variables}
\label{sec:2digits3var}

\begin{figure}[t]
\begin{center}
\centerline{\includegraphics[width=0.95\columnwidth]{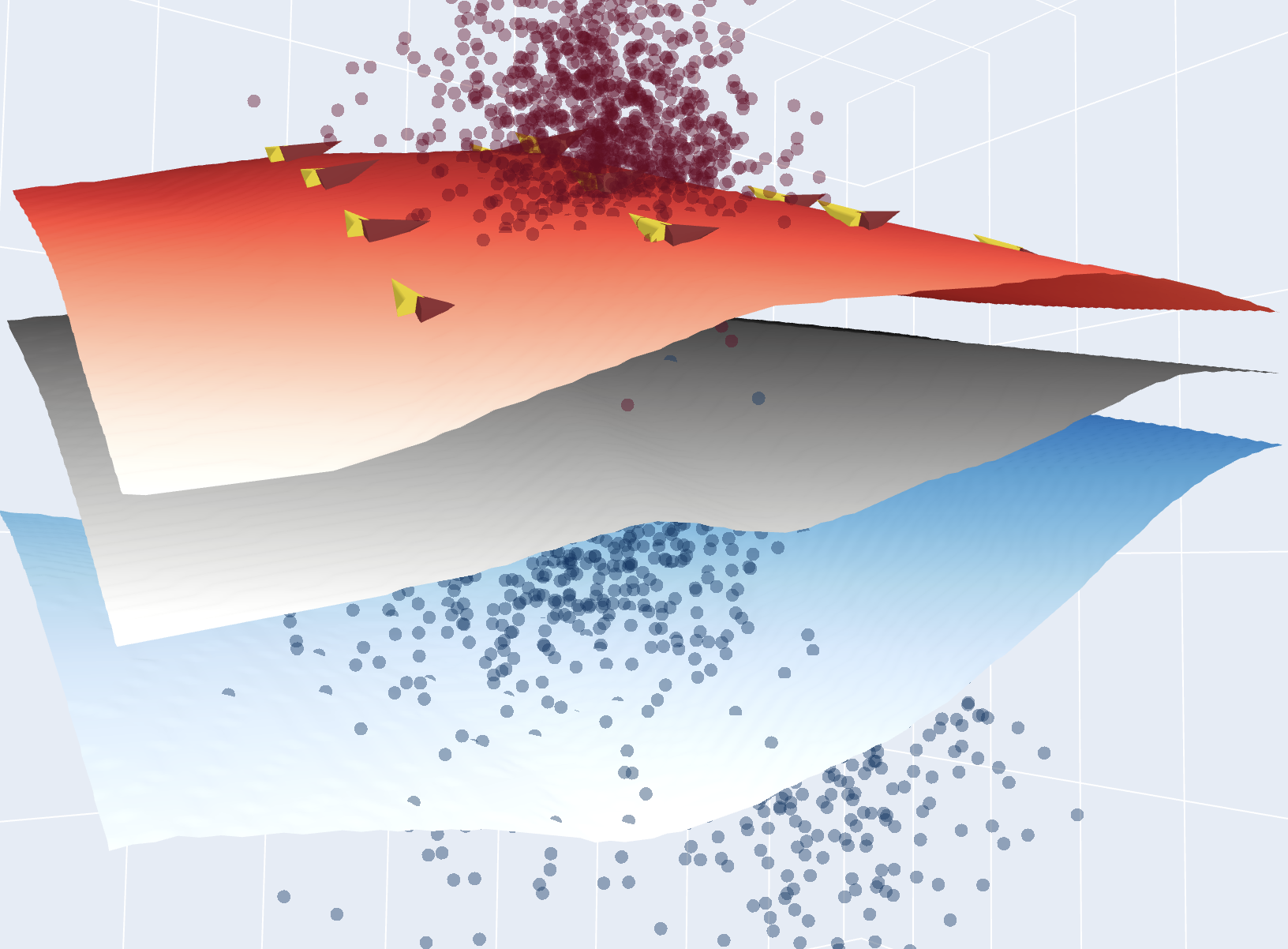}}
\caption{An illustration of the three-dimensional latent space of the example in Section~\ref{sec:2digits3var}. The red (blue) points represent validation images with true labels 0 and 1, respectively. The three surfaces show three level sets of the oracle $\psi$ with likelihood for ``zero" of 0.9999, 0.5 and 0.0001 (from top to bottom). The yellow and red arrows denote vectors sampled from the two latent flows found by our method (for simplicity, we only sample points on the upper surface). }
\label{fig:2v3d_vectors}
\end{center}
\end{figure}

In this subsection we expand the latent space of our toy example to $\ell=3$ dimensions, which, in contrast to the example in the previous subsection, will allow us to find a second, non-trivial, latent flow which generates a symmetry. The analysis proceeds as before, except the autoencoder is retrained with a three-dimensional bottleneck. The final result is presented in Figure~\ref{fig:2v3d_vectors}, which shows the three-dimensional latent space, together with the validation data which forms two clusters of zeros and ones. Then we superimpose three surfaces which are level sets of the oracle $\psi$ with likelihood for ``zero" of 0.9999, 0.5, and 0.0001, respectively. A generic local symmetry transformation is tangential to these surfaces, which is precisely the result we find when we train a NN for a single symmetry transformation $\mathbf g$.

The real power of our method lies in its ability to simultaneously find multiple orthogonal symmetry generators. Indeed, when we retrain for the case of $N_g=2$, we are able to find two orthogonal vector flows, as illustrated  in Figure~\ref{fig:2v3d_vectors} with the yellow and red arrows (for simplicity, we only show a small number of vectors sampled from the top surface). Note that all arrows are tangential to the level set surface, as expected for an invariant latent flow.


We chose this particular toy example because it represents the most complicated situation where there are multiple non-trivial generators, which can still be easily visualized. Once we generalize to higher dimensions in the next subsection, such a simple visualization will not be possible, but the results can be intuitively understood in a similar fashion.

\subsection{Ten Categories and $\ell=16$ Latent Variables}
\label{sec:symmetries}

\begin{figure}[t]
\begin{center}
\includegraphics[width=0.95\columnwidth]{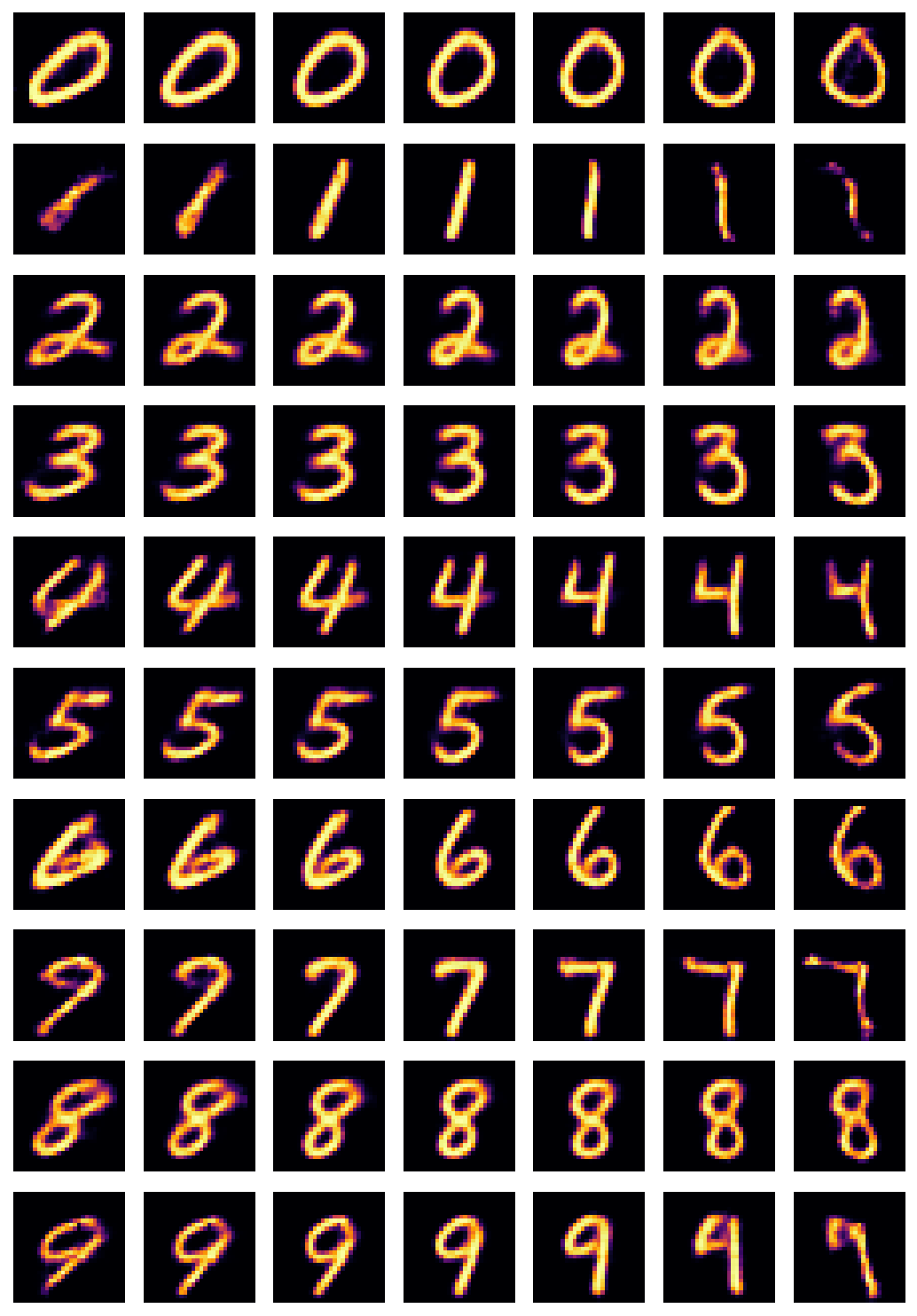}
\caption{Symmetric morphing of images along streamlines of the 16-dimensional latent flow found in Section~\ref{sec:symmetries}. The images in the middle column represent the ``platonic" digits in the dataset (the cluster centers for each of the ten classes). The remaining six images in each row are obtained by moving along or against the streamline passing through the respective platonic image. }
\label{fig:16v10d}
\end{center}
\end{figure}

In this subsection we present our final example in which we consider all ten classes of digits and use an autoencoder with an $\ell=16$ dimensional bottleneck. To improve performance, we found it useful to increase the number of mapping layers (i.e., those immediately before and after the bottleneck) from 1 to 3. Proceeding as before, we train a classifier $\vec\psi$ with 10 softmax outputs and a neural network for the symmetry transformation $\mathbf g$. The result is illustrated in Figure~\ref{fig:16v10d}, where in analogy to Figure~\ref{fig:2v2d_platonic} we show a series of decoded images along a streamline of the latent flow. The center images in Figure~\ref{fig:16v10d} represent the platonic digits in the dataset. From each of those ten starting points, we follow the respective streamline for 6000 steps of $\varepsilon=\pm 10^{-3}$. The three images to the left and to the right of the central one are obtained after 2000, 4000, and 6000 such steps in each direction. The images in each row are classified correctly with predicted probability 1.0. This validates our method for the case of a single symmetry flow. 

We also simultaneously trained multiple generators and found non-trivial orthogonal flows, even for relatively large numbers of generators ($N_g\sim 10$), which hints at the presence of non-Abelian symmetries. This study is only scratching the surface of a promising new research direction to reveal the rich symmetry structure of complex datasets, which we leave for future work. The end goal should be a rigorous understanding of their algebraic and topological properties.

\section{Summary and Conclusions}
\label{sec:conclusions}

In this paper, we studied a fundamental problem in data science which is commonly encountered in many fields: what is the symmetry of a labeled dataset, and how  does one identify its group structure? For this purpose, we applied a deep-learning method which models the generic symmetry transformation and its corresponding generators with a fully connected neural network. We trained the network with a specially constructed loss function ensuring the desired symmetry properties, namely: i) the transformed samples preserve the oracle output (invariance); ii) the transformations are non-trivial (normalization); iii) in case of multiple symmetries, the learned transformations must be distinct (orthogonality) and iv) must form an algebra (closure).

We leverage the fact that a non-trivial symmetry induces a corresponding flow in the latent space learned by an autoencoder. An important advantage of our approach is that we do not require any advance knowledge of what symmetries can be expected in the data, i.e., instead of testing for symmetries from a predefined list of possibilities, we learn the symmetry directly. 

We demonstrated the performance of our method on the standard MNIST digit dataset, showing that the learned transformations correspond to interpretable non-trivial transformations of the images. Future work could extend this approach to a much broader range of data and symmetry types.

\bibliography{example_paper}
\bibliographystyle{icml2023}




\end{document}